\documentclass[letterpaper, 10 pt, conference]{ieeeconf}
\IEEEoverridecommandlockouts
\usepackage{amsmath,amsfonts}
\usepackage{multirow}
\usepackage{algorithmic}
\usepackage{algorithm}
\usepackage{array}
\usepackage{caption}
\usepackage[caption=false,font=normalsize,labelfont=sf,textfont=sf]{subfig}
\usepackage{textcomp}
\usepackage{url}
\usepackage{verbatim}
\usepackage{graphicx}
\usepackage{cite}
\usepackage{booktabs}
\usepackage[english]{babel}
\newtheorem{problem}{Problem}
\DeclareMathOperator{\atantwo}{atan2}

\begin{document}

\title{Intent-Informed Model Predictive Control for Shared Autonomy in Remote Manipulation Tasks}

\author{Lorena Maria Genua$^{1,2}$, Camilla Tabasso$^{1}$, Mya Schroder$^{1,3}$, Joshua Migdal$^{1}$, Tye Brady$^{1}$, Ta\c{s}k{\i}n Pad{\i}r$^{1}$
\thanks{*This work was done when Lorena Maria Genua was a co-op at Amazon.}
\thanks{
$^\ddagger$Ta\c{s}k{\i}n Pad{\i}r holds concurrent appointments as a Professor of Electrical and Computer Engineering at Northeastern University and as an Amazon Scholar. This paper describes work performed at Amazon and is not associated with Northeastern University.
}%
\thanks{$^{1}$Amazon Fulfillment Technologies \& Robotics, Westborough MA, USA
        {\tt\footnotesize \{camtab, jmigdal, ptaskin\}@amazon.com}}%
\thanks{$^{2}$Institute for Experiential Robotics, Northeastern University, Boston MA, USA
        {\tt\footnotesize (genua.l@northeastern.edu)}}%
\thanks{$^{3}$School of Computer, Data \& Information Sciences, University of Wisconsin - Madison, Madison WI, USA
        {\tt\footnotesize (mcschroder@wisc.edu)}}%
}



\maketitle

\begin{abstract}
Shared autonomy is a control paradigm that combines human cognitive abilities with robot precision to enable effective remote manipulation in complex environments.
A key challenge is how to adapt the level of autonomy using information extracted from the human, task, and/or environment.
In this paper, we propose a novel arbitration strategy that dynamically adjusts the level of autonomy in the system using the entropy of a belief distribution derived from a multimodal intent inference.
Human intent is modeled with a Dynamic Bayesian Network (DBN), which updates beliefs over goals given real-time operator gaze data and robot arm motion during teleoperation.
We then formulate the blending of human input and autonomous control as a Model Predictive Control (MPC) problem, where the system's autonomous behavior is guided by the belief distribution.
This approach enables continuous, gradual transitions across the full spectrum of autonomy levels, from direct teleoperation to full autonomy.
To validate our approach, we conducted experiments to compare the performance of our proposed algorithm against two MPC-based shared autonomy methods.
Compared to the baselines, our approach achieved reductions of at least 61\% in object retrieval time, 44\% in trajectory length, and 60\% in human control input.
Participants reported high perceived control, predictability, intent recognition, and responsiveness, along with low workload.

\end{abstract}
\vspace{2ex}
\section{Introduction}
Teleoperation has been effective for controlling robots in remote and hazardous environments.
However, the full control of robot manipulators poses several challenges due to the reduced dexterity in precise manipulation, limited situational awareness from camera views, and communication delays~\cite{Ruixing:camera_viewpoint, Jean:latency}.
Recent advances in teleoperation systems, such as those developed by teams participating in the ANA Avatar XPRIZE~\cite{Schwarz:XPrize, Luo:XPrize, Hauser:XPrize}, have addressed these issues by integrating virtual reality interfaces with multisensory feedback to enable high-dexterity manipulation in remote environments.
While these systems enhance operator embodiment, the human operator is responsible for continuous low-level control and decision-making during task execution.
As a result, prolonged operation leads to increased operator fatigue and degraded performance~\cite{YiTing:workload}.
Conversely, fully autonomous robots fall short in complex and ambiguous situations found in unpredictable and unstructured environments, often necessitating human intervention~\cite{Devleena:explanable_failures}.
Shared autonomy enables the collaborative control between human operators and autonomous systems~\cite{Losey:shared_control_review}, therefore leveraging human cognitive abilities for high-level decision-making and robotic precision for low-level control.
\begin{figure}[t!]
    \centering
    \includegraphics[width=\linewidth]{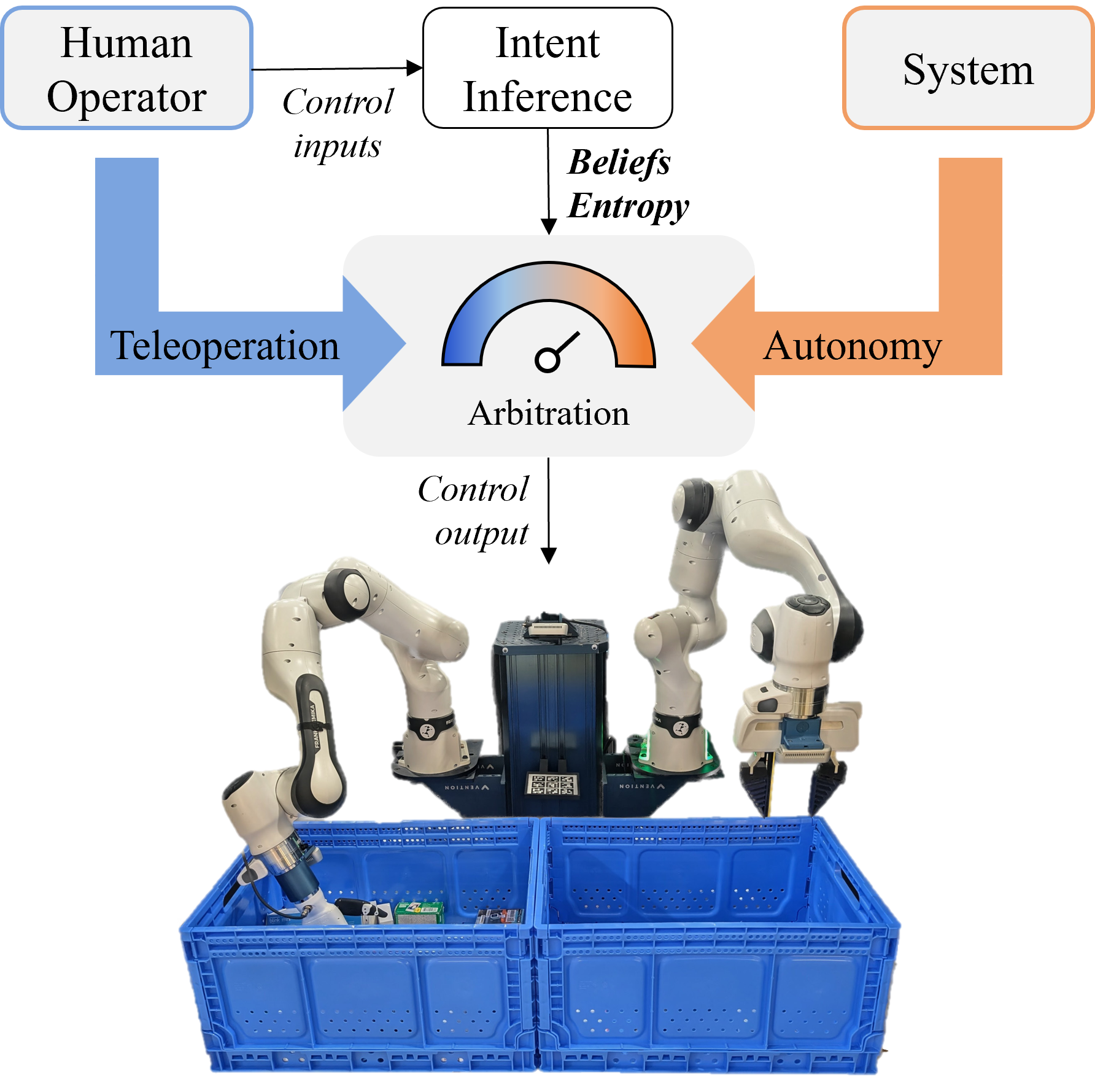}
    \caption{\textbf{Intent-Informed Model Predictive Control} uses the \textbf{entropy} of the belief distribution derived from human intent inference to determine the \textbf{control authority} between the human operator and the system.}
    \label{fig:system-setup}
\end{figure}
Two key aspects of shared autonomy are: arbitration, which is who to allocate control authority based on different contextual factors, and policy blending, how to combine human input with autonomous actions.
Existing related work has explored various arbitration strategies for shared-autonomy based on performance metrics and user-centric approaches~\cite{Saeidi:arbitration_trust, Zhu:arbitration_gaussian, Oh:arbitration_subpolicies, Xing:arbitration_fuzzy_logic, Milliken:arbitration_user_expertise, Li:arbitration_multisource_uncertainty, Balachandran:arbitration_bayesian_filter, Motaharifar:arbitration_training_performance}.
However, how to smoothly adjust the autonomy levels by leveraging the information extracted from the human still remains an open research question~\cite{Gaofeng:shared_autonomy_review}.
The existing works do not fully leverage real-time human intent inference from multimodal inputs to inform the arbitration process.
\\In this work, we propose an Intent-Informed Model Predictive Control (MPC) framework for adaptive shared autonomy in remote manipulation tasks.
Our approach combines human intent inference using a Dynamic Bayesian Network with MPC to dynamically adjust the level of autonomy based on the confidence in the inferred operator intent.
\\Our main contributions are the following:

1) We introduce a \textbf{novel arbitration strategy} that uses the \textbf{entropy of belief distributions} as a measure of intent inference uncertainty. This allows a gradual and continuous autonomy-level adaptation, shifting control authority between direct teleoperation and full autonomy.

2) We formulate the \textbf{policy blending} problem as a \textbf{Model Predictive Control (MPC) problem}, where the probabilities from the intent inference guide the autonomous control.

3) We conducted experimental studies with 20 participants and compared our approach with two existing MPC formulations for shared-autonomy.

\section{Related work}
\subsection{Arbitration strategies}
Arbitration determines how the control authority is allocated between a human operator and an autonomous agent.
Early approaches used a discrete switching between manual and autonomous control modes, often triggered by intent confidence~\cite{Oh:arbitration_subpolicies, jabbour:methodB}.
Such methods typically result in abrupt transitions, which can be disorienting for users and may fail to optimize task performance.
More recent work has explored continuous arbitration strategies, which allow smooth adjustments across the spectrum of autonomy levels.
Various methods for adjusting arbitration weights have been proposed, focusing on different performance metrics. 
For instance, Xing et al.~\cite{Xing:arbitration_fuzzy_logic} developed a fuzzy logic-based arbitration which modulates control weights according to interaction force, human-robot velocity, and proximity to obstacles.
Balachandran et al.~\cite{Balachandran:arbitration_bayesian_filter} apply Bayesian filtering to estimate manipulation performance and allocate control accordingly.
However, because these methods do not incorporate intent inference, they may lead to conflicts between human intentions and the actions of the autonomous controller.
Other works have addressed this by relying on human control input and distance to targets~\cite{Zhu:arbitration_gaussian, Lima:methodA}.
These approaches struggle in cluttered environments and are challenging for novice teleoperation users, who tend to exhibit jerkier manipulation control and experience sharp declines in performance as task complexity increases~\cite{Tugal:teleoperation_novice}. 
User-centric approaches, conversely, adjust the level of autonomy based on user expertise and training experience~\cite{Milliken:arbitration_user_expertise, Motaharifar:arbitration_training_performance}, or models of bilateral trust between humans and robots~\cite{Saeidi:arbitration_trust}.
\\The challenge of accurately anticipating human intentions through multimodal information integration, and use it to adapt the level of autonomy, still remains an open research question~\cite{Gaofeng:shared_autonomy_review}.
Existing approaches have overlooked the potential of biological signals for intent inference.
Human gaze, in particular, provides valuable insight into task-relevant attention and intended actions~\cite{Land:gaze}, yet it has not been leveraged for arbitration strategies.
In this work, we propose an intent inference method that incorporates multimodal inputs, e.g. gaze tracking and motion direction.
We then use the entropy of the beliefs, a measure of uncertainty in the system, as an arbitration strategy to adjust the level of autonomy.

\subsection{Model Predictive Control for Shared Autonomy} \label{related_work:mpc}
Model Predictive Control (MPC) is an optimal control technique that uses a model of the system to predict the future behavior, then optimizes the control inputs to minimize a desired cost function.
In the context of shared autonomy, the application of MPC remains mostly limited to trajectory correction and obstacle avoidance.
Several works have used MPC to satisfy system constraints and prevent collisions with static and dynamic obstacles while following human input~\cite{Rubagotti:shared_control_avoidance, Saparia:shared_control_avoidance, Storms:shared_control_avoidance, Hu:shared_control_avoidance}.
More recently, two works have explored MPC as a blending strategy, which we use as comparison methods in this paper.
Lima et al.~\cite{Lima:methodA} provide continuous assistance towards multiple targets using distance-based weighting in their cost function to switch between autonomous motion and user input.
While this approach blends user input with robot autonomy, it has two main limitations. 
First, it does not provide a gradual transition between autonomy levels, resulting in abrupt changes in control.
Second, because it relies on distance to targets, it faces challenges in cluttered environments where objects are in close proximity.
\\Jabbour et al.~\cite{jabbour:methodB} use distance to the target as a weight to distribute the control authority.
This method however, requires a predefined goal location.
We include this work for comparison as it combines a common shared autonomy strategy (proximity-based assistance) with MPC formulation.
While common in shared autonomy, distance-based thresholds methods do not provide a continuous blending and require predetermined goals.
Our work extends the use of MPC in shared autonomy by introducing two novelties.
First, we employ MPC as a continuous blending strategy, enabling a gradual transition in control authority from full teleoperation to full autonomy.
Second, we incorporate the entropy of intent inference beliefs as a weight to dynamically adjust the level of autonomy in the system.
Additionally, our formulation includes both position and orientation tracking, unlike the previous two approaches which focus solely on position.

\section{Problem statement}
\begin{figure*}[t!]
    \centering
    \includegraphics[width=\textwidth]{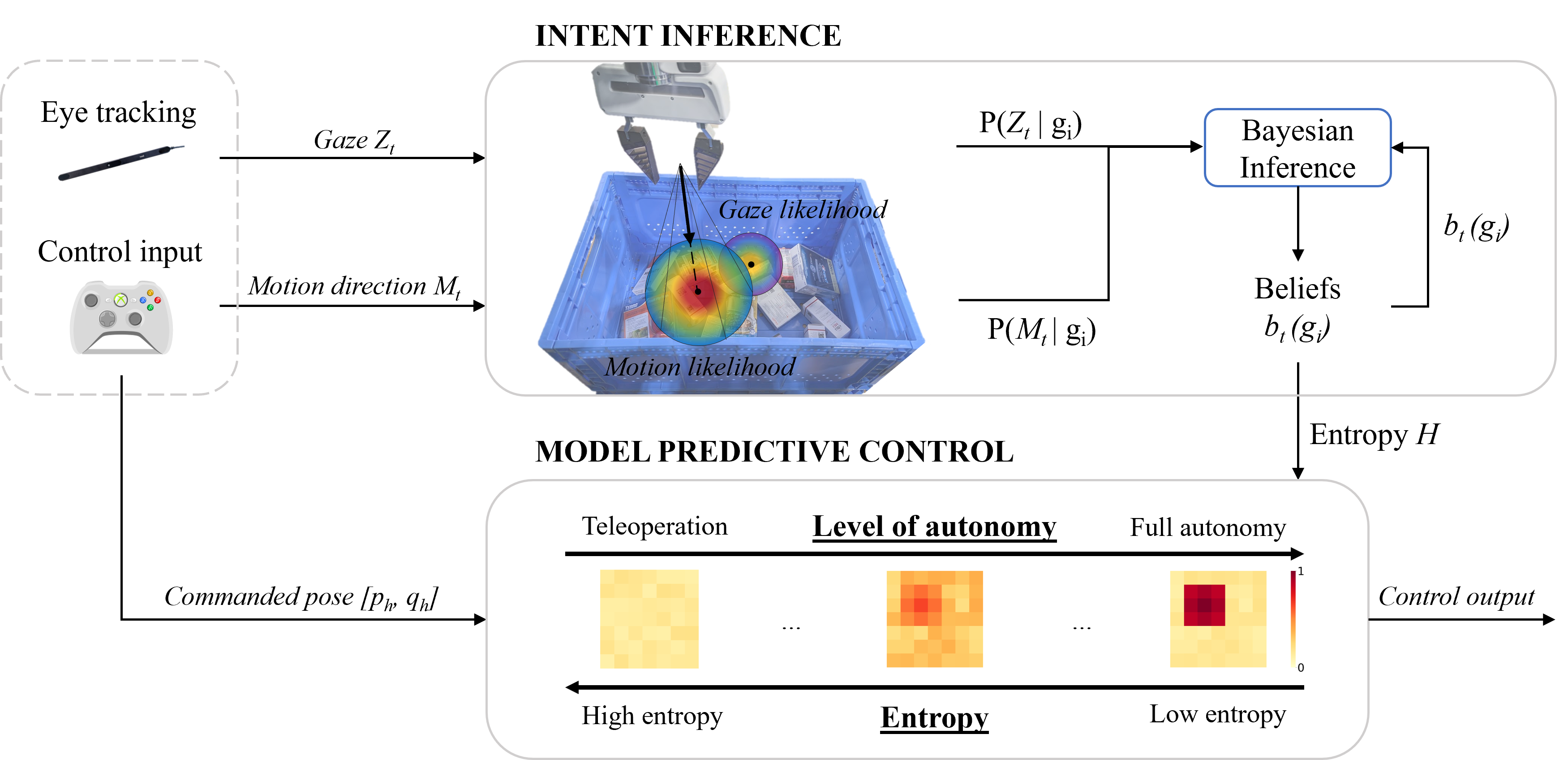}
    \caption{\textbf{Methodology overview.} Our approach combines intent inference using a Dynamic Bayesian Network (DBN) with Model Predictive Control (MPC) for adaptive shared autonomy. The DBN continuously updates beliefs on the goals based on operator gaze and robot motion. The MPC formulation uses the entropy of these beliefs to adjust the level of autonomy, ranging from direct teleoperation when entropy is high to full autonomy when entropy is low.}
    \label{fig:methodology}
\end{figure*}
Consider a finite set of $n_g$ discrete goals $\mathcal{G} = \{g_1, g_2, \ldots, g_n\}$, where $g^* \in \mathcal{G}$ represents the desired goal of the human operator.
The robot does not have a prior knowledge of $g^*$ and must assist the human to pick the desired goal.
Then, the objective of this work is twofold: first, we perform human intent inference to determine the most likely target that the operator wants to retrieve.
Secondly, we develop a control algorithms which increases the amount of autonomy in the system in a manner that is proportional to the confidence of the inference.
This results in a continuous, bidirectional reallocation of control authority between the operator and the autonomous controller, driven by changes in intent inference uncertainty.

\section{Intent Inference}

We formulate intent inference as a Bayesian filtering problem within a Dynamic Bayesian Network (DBN), modeling the human's intent as a probability distribution over a set of known goals given a sequence of observations.
Let $\mathcal{G} = \{g_1, g_2, \dots, g_{n_g}\}$ denote the set of possible goals, and let $\theta_t = [\theta_t^1, \theta_t^2, \dots, \theta_t^j]$ represent the vector of $j$ observed features at time $t$.
In this work, $j=2$, with $\theta_t^1 = M_t$ representing the directional control input and $\theta_t^2 = Z_t$ representing the human gaze.

\subsection{Gaze filtering}
Human gaze provides a valuable cue for intent inference, as people tend to direct their visual attention toward task-relevant objects~\cite{Land:gaze}.
We use the Identification by Velocity Threshold (I-VT) algorithm~\cite{salvucci:algorithms-fixations} to differentiate raw gaze data into fixations (stable visual attention) and saccades (rapid transitions). 
During saccades, visual sensitivity is functionally suppressed (known as saccadic masking), making these data points irrelevant for intent inference.
Point-to-point angular velocity is computed between consecutive gaze samples.
All samples with angular velocity below $30^\circ$/s and duration of at least 100 ms are labeled as fixations, and are included in $Z_t$.
%
%

\subsection{Bayesian Update Rule}
Intention recognition is commonly addressed using probabilistic graphical models, such as Hidden Markov Models (HMMs) and Dynamic Bayesian Networks (DBNs), which infer a latent intent state from a sequence of noisy observations~\cite{Hoffman2024}.
In this work, we model the operator’s intent as a discrete latent variable over a finite set of candidate goals and estimate it using a Dynamic Bayesian Network.
This formulation allows us to recursively fuse multimodal information (motion control commands and gaze) into a belief distribution that is updated online during teleoperation.
The posterior probability of goal $g_i$ at time $t$ given all observations up to that time is computed via Bayes’ rule:
\begin{equation}
P(g_i \mid \theta_{1:t}) = \frac{P(\theta_t \mid g_i, \theta_{1:t-1}) \, P(g_i \mid \theta_{1:t-1})}{P(\theta_t \mid \theta_{1:t-1})}.
\end{equation}
Applying the Markov assumption of conditional independence of the current observation from past observations given the goal $g_i$, we can simplify $P(\theta_t \mid g_i, \theta_{1:t-1})$ as $ P(\theta_t \mid g_i)$.
We also assume the observation sources to be conditionally independent of each other, such that
$P(\theta_t \mid g_i) = P(M_t \mid g_i) \, P(Z_t \mid g_i)$.
Therefore, the belief of goal $g_i$ at time $t$ can be written as:
\begin{equation}
b_t(g_i) \propto P(M_t \mid g_i) \, P(Z_t \mid g_i) \, b_{t-1}(g_i),
\end{equation}
where $b_{t-1}(g_i) = P(g_i \mid \theta_{t-1})$ is the prior belief from the previous time step.
To reduce sensitivity to noisy observations, we introduce a smoothing factor $\alpha$:
\begin{align}
b_t(g_i) &= \alpha \, b_{t-1}(g_i) \nonumber \\
        &+ (1-\alpha)\, \big[b_{t-1}(g_i)\, P(M_t|g_i) \, P(Z_t|g_i) \big]
\end{align}

Here:
\begin{itemize}
    \item $\alpha \in [0,1]$ controls the rate of belief convergence and the system's sensitivity to new information,
    \item $P(M_t|g_i)$ and $P(Z_t|g_i)$ are Gaussian likelihoods of observed motion and gaze, respectively.
\end{itemize}
 The prior belief is initialized as a uniform distribution \begin{equation}
P(g_i|\mathcal{\theta}_0) = \frac{1}{n_g}, \quad \forall g_i \in \mathcal{G}.
\end{equation}

\subsection{Entropy}
Given the posterior probability $b_t(g_i)$, we calculate the entropy of the belief distribution $H$ using Shannon's equation~\cite{Shannon:entropy}:
\begin{equation}
H = -\sum_{i=1}^{n_g} b_t(g_i) \log b_t(g_i)
\label{eq:entropy}
\end{equation}
where $n_g$ is the number of goals in $\mathcal{G}$.
Shannon entropy summarizes the dispersion of the belief distribution: it is near zero when one goal dominates (high intent confidence) and is maximized when beliefs are uniform (complete uncertainty).
$H$ is then normalized in order to obtain a standardized measure of uncertainty that is independent of the number of goals. 
\begin{equation}
\hat{H} = \frac{H}{H_{max}}
\label{eq:normalized_entropy}
\end{equation}

where $H_{max}$ is the maximum possible entropy for the given number of goals.
The maximum entropy occurs when all goals have a uniform distribution and is calculated as:
\begin{equation}
H_{max} = \log(n_g)
\label{eq:max_entropy}
\end{equation}
$\hat{H}$ quantifies the uncertainty in the intent inference and will be used in the optimization problem as a weight to adjust the autonomy level.
Because $\hat{H}\in[0,1]$ varies smoothly with the belief distribution, it enables continuous and gradual arbitration without discrete switching or heuristic thresholds.
\section{Model predictive control}
In this section, we present the mathematical formulation of the assisted teleoperation problem.

\subsection{Dynamic model}
In order to reduce the computational complexity and achieve real time performance, we adopt a single integrator system to model the dynamics. 
This simplification is possible because low-level joint dynamics and control are handled by the robot’s internal controllers. 
\\Specifically, consider the state vector at time $t_k$, $x_k = [p_k, q_k]^\top \in \mathbb{R}^7$, where $p_k = [p_{x,k}, p_{y,k}, p_{z,k}]^\top$ and $q_k = [q_{w,k}, q_{x,k}, q_{y,k}, q_{z,k}]^\top$ represent the position and orientation of the end effector, respectively. Furthermore, let the input vector at time $t_k$ be defined as $u_k = [v_k, \omega_k]^\top \in \mathbb{R}^6$ where $v_k = [v_{x,k}, v_{y,k}, v_{z,k}]^\top$ and $\omega_k = [0, \phi_k, \theta_k, \psi_k]^\top$ are the linear and angular velocity components, respectively. Then, the dynamic model is defined as

\begin{equation}
    \begin{bmatrix}
        \mathbf{p}_{k+1} \\
        \mathbf{q}_{k+1}
    \end{bmatrix} =
    \begin{bmatrix}
        \mathbf{p}_k \\
        \mathbf{q}_k
    \end{bmatrix} + \Delta t
    \begin{bmatrix}
        \mathbf{v}_k \\
        \frac{1}{2} \mathbf{q}_k \circ \mathbf{\omega}_k
    \end{bmatrix}
\end{equation}

\noindent where $\Delta t$ is the time step.
The term $\frac{1}{2}\mathbf{q}_k \circ \mathbf{\omega}_k$ represents the rate of change of the orientation quaternion, and $\circ$ defines the quaternion product found as:
\begin{equation}
\mathbf{q}_k \circ {\omega}_k
=
\begin{bmatrix}
- \mathbf{q}_{v,k} {\omega}_{v,k} \\
q_{\omega,k} {\omega}_{v,k} + \mathbf{q}_{v,k} \times {\omega}_{v,k}
\end{bmatrix}
\end{equation}
where $\mathbf{q}_{v,k} = [q_{x,k}, q_{y,k}, q_{z,k}]^\top$ is the vector part of the orientation quaternion, and $\mathbf{\omega}_{v,k} = [\phi_k, \theta_k, \psi_k]^\top$ is the vector part of the angular velocity quaternion.

\subsection{MPC formulation}

One of the main contributions of our work is the formulation of a novel policy blending strategy found by solving an MPC problem. MPC enables the system to determine the optimal level of autonomy based on the confidence of the intent inference, while simultaneously enforcing constraints satisfaction to ensure feasibility. The MPC problem is defined as follows.

\begin{problem}
Let the operator's commanded position and orientation at time $t_k$ be $p_{h,k}$ and $q_{h,k}$, respectively. Furthermore, let $P$ and $Q$ be diagonal positive definite weight matrices and $w_{rot}$ be a positive control gain.
\\Then, for the time horizon $T$, the assisted teleoperation problem is defined as follows:

\begin{align}
    \min_{x, u} & \sum_{k=1}^{T} \Big[\hat{H} \| p_k - p_{h,k} \|_{Q} \nonumber \\
     &+ 2 w_{rot} \atantwo \Big(\sqrt{\hat{q}_{x,k}^2 + \hat{q}_{y,k}^2 + \hat{q}_{z,k}^2}, \hat{q}_{w,k}^2 \Big) \nonumber \\
    &+ (1-\hat{H}) \sum_{i=1}^{n_g} b_t(g_i) \| p_k - g_{i} \|_{P} \Big]
\end{align}
\hspace{10mm} s.t.
\begin{align}
    p_{z,k} \geq p_{z, \min}, \\
    || p_{k} - p_{o, k} || \geq d_{\min}, \\
    q_{\min} \leq q_k \leq q_{\max}, \\
    u_{\min} \leq u_k \leq u_{\max},
\end{align}
\noindent where $\hat{q} = q_{h,k} \cdot q_{k}^{-1}$ is the error quaternion, $p_{o,k}$ is the position of the obstacles at time $t_k$. Note that the notation $||\cdot||_{Q}$ represent the quadratic form defined as $(\cdot)^T Q (\cdot)$.
\end{problem}
The first two terms in the cost function relate to tracking the human position and orientation inputs, respectively. Conversely, the last term incentivizes the robot to move towards the estimated human goal predicted by the intent inference algorithm.
The novelty of our formulation is the use of the entropy of the system, $\hat{H}$, as a weight.
The intuition behind this is that as the confidence of the intent inference increases, the entropy will vanish.
This results in a high level of autonomy. Conversely, when the entropy is high, which corresponds to low confidence, the last term in the cost function corresponding to full teleoperation will be maximized.
\\The resulting nonlinear optimization problem can be solved with any off the shelf solver without any significant changes in the performance of our framework.

\section{Experiments}
To evaluate the efficacy of our proposed method, we conducted experiments with N=20 participants (11 males and 9 females, 22.45 $\pm$ 3.48 years).
The parameters used for Intent-Informed MPC were determined experimentally and are described in Table~\ref{tab:parameters}. The nonlinear optimization problem is solved using Sequential Least Squares Programming (SLSQP) implemented with SciPy~\cite{2020SciPy}.
%
\begin{table}[h]
\centering
\caption{System Parameters}
\vspace{1ex}
\renewcommand{\arraystretch}{1.4} 
\begin{tabular}{*{6}{>{\centering\arraybackslash}p{0.11\linewidth}}}
\textit{Parameter} & \textit{Value} & \textit{Parameter} & \textit{Value} \\
\hline
$\alpha$ & 0.2 & $P$ & $30I$\\
$T$ & 10 & $Q$ & $30I$ \\
$n_{g}$ & $6$ & $w_{rot}$ & $5$ \\
\hline
\end{tabular}
\label{tab:parameters}
\end{table}
\\\textbf{Baselines:}
Our approach was benchmarked against two methods that frame shared autonomy as a Model Predictive Control (MPC) problem.
These methods were selected because they are state-of-the-art MPC formulations for shared autonomy but use a different arbitration strategy (on distance- and proximity-based arbitration).
The first comparative method, here referred to as \textbf{A-MPC}~\cite{Lima:methodA}, presents a continuous policy adaptation towards multiple goals using distance-based weighting.
The control parameters were tuned for our system as they were not specified in the original paper.
The second method under comparison~\cite{jabbour:methodB}, implements an arbitration technique where the authority level shifts from full assistance to full human control based on the proximity to the goal.
This method will be referred to as \textbf{B-MPC}.
The same parameters specified in the paper were used for our system.
Our proposed approach differs by providing continuous, gradual autonomy-level adaptation since the beginning of the task, and it's responsive to changes in human intent.
Because of that, we hypothesize that Intent-Informed MPC will:
\textit{H1) Reduce object retrieval time}, \textit{H2) Lower human control input}, and \textit{H3) Reduce trajectory length.}
\begin{figure}[h]
    \centering
    \includegraphics[width=\linewidth]{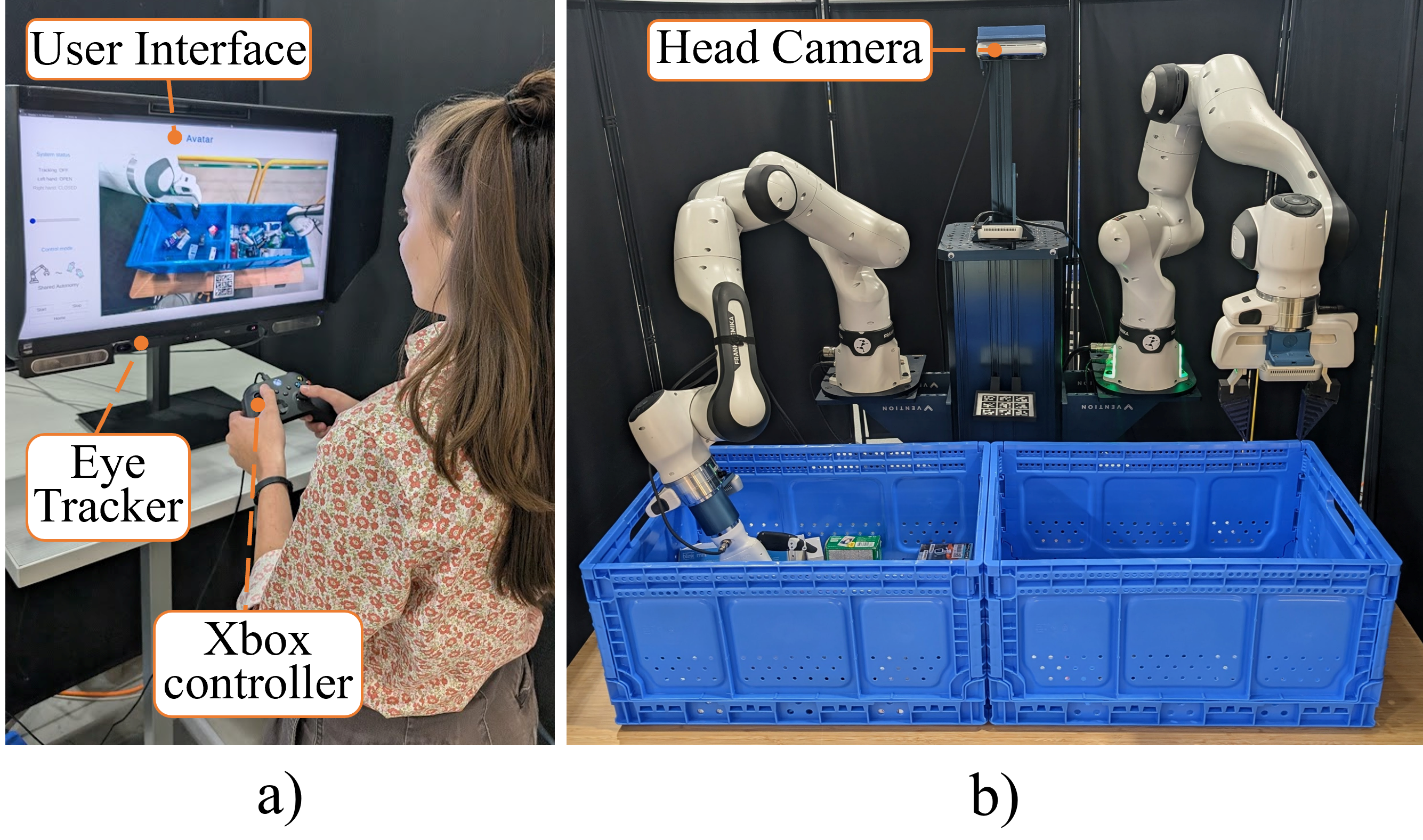}
    \caption{\textbf{Experimental setup.} a) The operator uses an Xbox controller to teleoperate the robot through a screen-based user interface. The eye tracker is mounted on the monitor. b) Robotic system. For the experiment, we use the left arm with the FinRay gripper. The right arm is not operational.}
    \label{fig:system-setup}
\end{figure}
\\\textbf{Task and Experiment Setup:}
Fig.~\ref{fig:system-setup} b) shows our experimental setup, which consisted of two Franka Research 3 robotic arms with 7 degrees of freedom (DoF).
For this experiment, only the left arm equipped with a FinRay gripper was utilized, while the right arm remained non-operational.
Participants teleoperated the robot with an Xbox controller to perform a tote-emptying task (Fig.~\ref{fig:system-setup} a)).
Joystick inputs were interpreted as cartesian motion commands for the end-effector.
The right joystick controlled planar motion, with lateral and longitudinal deflections mapped to velocity commands along the $x$ and $y$ axes of the robot base frame.
The left joystick controlled vertical motion, with forward and backward deflections mapped to velocity commands along the $z$ axis.
These velocity commands were obtained by scaling normalized joystick inputs with fixed gains and integrating them over time to update the desired cartesian end-effector position $p_{h,k}$.
The same control mapping was used across all modes.
To match the control modality of the baselines, we constrained our approach to position control only.
Participants used a screen-based interface developed with Unity Engine, which displayed the head camera view.
The monitor was equipped with a Tobii Pro Spark to track human gaze on the interface.
The task required participants to pick up three items from a tote in a specific order. After each item was picked up, the robot would automatically place it in the tote on the right.
\begin{figure*}[t!]
    \centering
    \includegraphics[width=\textwidth]{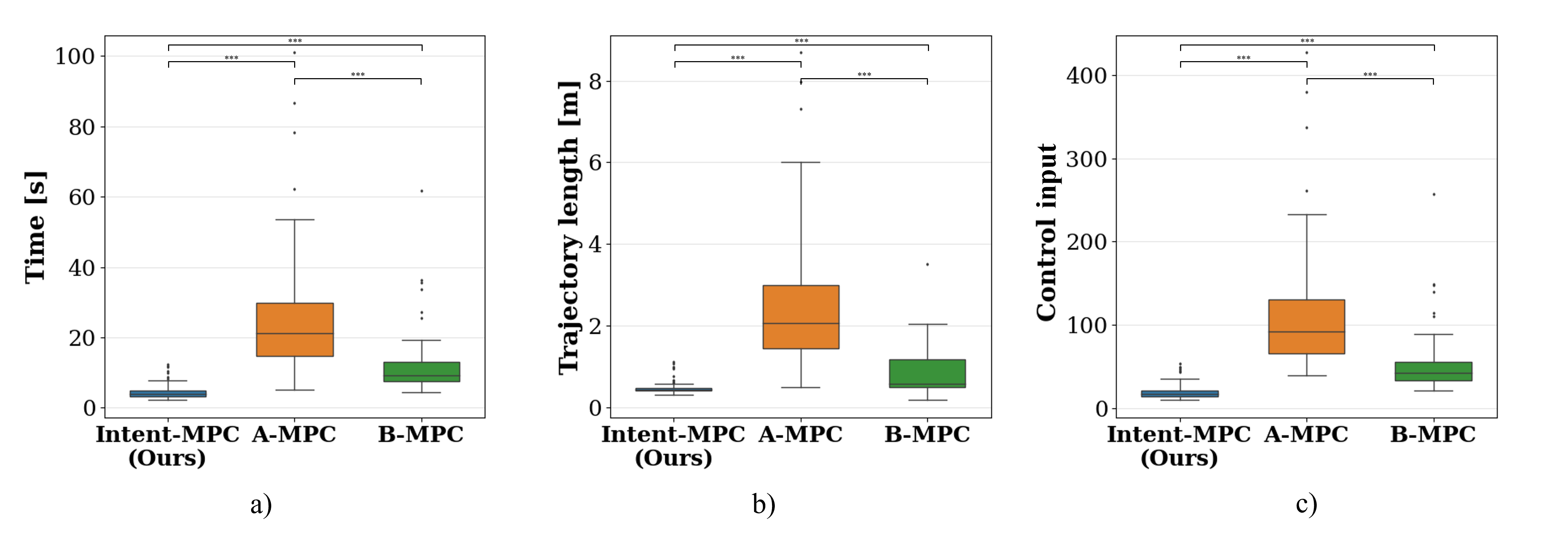}
    \caption{\textbf{Experimental results.} Our proposed method Intent-Informed MPC resulted in a) faster object retrieval time, b) lower trajectory length and c) lower human control input. Statistical significance ($p < 0.001$) is denoted by ***.}
    \label{fig:Results}
\end{figure*}
\\\textbf{Experiment Procedure and Data Collection:}
During the training phase, participants were instructed on how to teleoperate the robot and explained the different modalities.
They practiced each control mode on the same task used in the evaluation until they reported feeling comfortable using the system.
The evaluation phase consisted of two trials per mode, with each trial involving picking up three items from a tote in a specific sequence.
The order of modes was randomized, while the item sequence remained the same across participants.
Performance metrics were recorded for each individual item pick. 
Following each mode, participants completed the NASA Task Load Index (NASA-TLX) survey and a custom questionnaire, both expressed on a 7-point Likert scale.
For the custom questionnaire, they rated their agreement with the following statements:

\textit{Q1) I felt I was in control of the robot.}

\textit{Q2) The robot did what I expected it to do.}

\textit{Q3) The robot understood my intentions.}

\textit{Q4) The robot was responsive to my commands.}
\section{Results}
The non-parametric Friedman test was used to analyze the differences between the three modes, because the data didn't meet the assumptions of normality.
When significant, post-hoc pairwise Wilcoxon signed-rank test with Bonferroni correction was performed.
\subsection{Objective metrics}
All of our hypotheses were fully supported, with a significant difference across modes \textit{($p<0.001$)}.
Pairwise comparisons further revealed significant differences between all modes for each of the three metrics \textit{($p < 0.001$)}.

\vspace{1ex}
\textbf{\textit{H1) Intent-Informed MPC will reduce retreival time}.} 

The retreival time (Fig~\ref{fig:Results} a)), defined as the time taken to pick each individual item, was significantly lower for our proposed Intent-Informed MPC, with participants completing the task in an average of 4.40 $\pm$ 1.92 seconds.
This was faster than B-MPC, which required 11.22 $\pm$ 7.26 seconds, while A-MPC had the slowest performance (24.70 $\pm$ 15.47 seconds).

\vspace{1ex}
\textbf{\textit{H2) Intent-Informed MPC will reduce trajectory length}. 
}
The trajectory length (Fig~\ref{fig:Results} b)), representing the total distance traveled by the end-effector during task execution,  was significantly shorter for Intent-Informed MPC at 0.47 $\pm$ 0.13 meters, compared to B-MPC (0.84 $\pm$ 0.48 meters), and A-MPC (2.37 $\pm$ 1.42 meters).

\vspace{1ex}
\textbf{\textit{H3) Intent-Informed MPC will lower human control input}. 
}
We also analyzed the amount of human control input (Fig.~\ref{fig:Results} c)) during task execution.
It was calculated by summing the absolute values of all joystick axes inputs over time.
Intent-Informed MPC needed the least control input from operators, with an average of 19.70 $\pm$ 8.41. B-MPC required moderate input (49.76 $\pm$ 29.70) while A-MPC had the highest control input (108.12 $\pm$ 65.18).
\vspace{1ex}
\begin{figure}[t!]
    \centering
    \includegraphics[width=\linewidth]{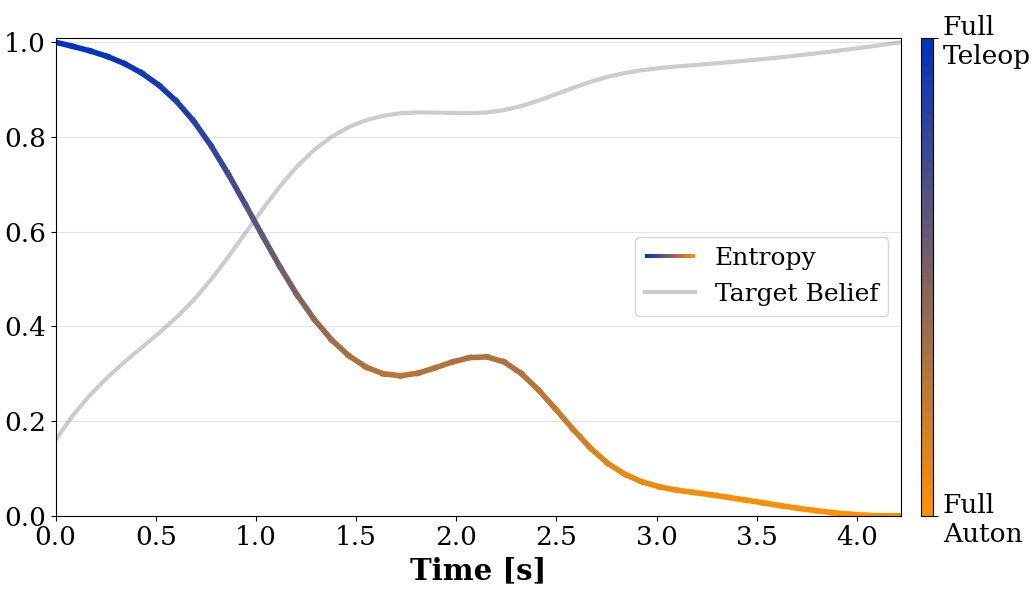}
    \caption{Time plot of entropy and target belief in a representative trial. As entropy decreases, the system increases the level of autonomy.}
    \label{fig:entropy}
\end{figure}
\begin{table*}[t]
\centering
\renewcommand{\arraystretch}{1.1}
\begin{tabular*}{\textwidth}{@{\extracolsep{\fill}}lcccccccccccc@{}}
\hline
\multirow{2}{*}{\centering \textbf{Question}}
& \multicolumn{2}{c}{\textbf{Intent-MPC (Ours)}} 
& \multicolumn{2}{c}{\textbf{A-MPC}} 
& \multicolumn{2}{c}{\textbf{B-MPC}} 
& \textbf{Intent vs A} 
& \textbf{Intent vs B} 
& \textbf{B vs A} \\
& Mean & SD & Mean & SD & Mean & SD 
& $p$-value & $p$-value & $p$-value \\
\hline
\textbf{Q1} Control
& 5.65 & 1.27 & 3.10 & 1.37 & 5.80 & 1.01
& {\boldmath$<0.01$} & $>0.05$ & {\boldmath$<0.001$} \\

\textbf{Q2} Predictability
& 6.35 & 0.75 & 2.80 & 1.28 & 6.35 & 0.67
& {\boldmath$<0.001$} & $>0.05$ & {\boldmath$<0.001$} \\

\textbf{Q3} Intent Understanding
& 6.35 & 0.99 & 3.00 & 1.34 & 6.10 & 1.21
& {\boldmath$<0.001$} & $>0.05$ & {\boldmath$<0.01$} \\

\textbf{Q4} Responsiveness
& 6.35 & 0.75 & 3.40 & 1.10 & 6.30 & 0.80
& {\boldmath$<0.01$} & $>0.05$ & {\boldmath$<0.01$} \\

\midrule
\textbf{Q5} Mental Demand
& 1.95 & 0.94 & 4.15 & 1.27 & 2.85 & 1.31
& {\boldmath$<0.01$} & $>0.05$ & $>0.05$ \\

\textbf{Q6} Physical Demand
& 1.35 & 0.59 & 2.55 & 2.01 & 1.90 & 1.17
& $>0.05$ & $>0.05$ & $>0.05$ \\

\textbf{Q7} Temporal Demand
& 2.45 & 1.64 & 3.15 & 1.39 & 2.35 & 1.42
& $>0.05$ & $>0.05$ & $>0.05$ \\

\textbf{Q8} Performance
& 6.40 & 0.68 & 4.05 & 1.39 & 6.55 & 0.51
& {\boldmath$<0.01$} & $>0.05$ & {\boldmath$<0.01$} \\

\textbf{Q9} Effort
& 1.95 & 0.89 & 4.50 & 1.32 & 2.95 & 1.73
& {\boldmath$<0.01$} & $>0.05$ & {\boldmath$<0.05$} \\

\textbf{Q10} Frustration
& 1.40 & 0.50 & 3.70 & 2.13 & 1.65 & 0.99
& {\boldmath$<0.05$} & $>0.05$ & {\boldmath$<0.01$} \\
\hline
\end{tabular*}
\caption{Mean and standard deviation (7-point Likert scale) for each subjective question across control modes. Pairwise comparisons report Bonferroni-corrected $p$-values with significant differences ($p<0.05$) shown in bold.}
\label{tab:subjective_full}
\end{table*}
Intent-Informed MPC outperforms both comparison methods in retreival time, trajectory length and control input.
This performance can be attributed to two factors: the continuous autonomy-levels adaption, and the efficiency of the intent inference.
The intent inference reached 90\% confidence in a target in an average of 2.60 $\pm$ 1.88 seconds across trials (as illustrated in Fig.~\ref{fig:entropy}).
Regarding autonomy level, Intent-Informed MPC had an average of 0.68 $\pm$ 0.10 \%.
Fig.~\ref{fig:entropy} visualizes this performance, showing the inverse relationship between the system's entropy and its level of autonomy.
The system is initialized with a uniform distribution and maximum entropy (teleoperation state).
As the confidence in the target belief increases and the entropy decreases, the system’s autonomy level smoothly transitions from teleoperation to full autonomy.
This continuous adaption of autonomy levels explains the significant difference with B-MPC  \textit{($p<0.0001$)}, which had an average autonomy level of 0.23 $\pm$ 0.09 \%.
It's worth noting that A-MPC was excluded from this comparison due to the absence of an explicit arbitration variable in their formulation.
\subsection{Subjective metrics}
Results for the custom questionnaire (Q1-Q4) and NASA-TLX (Q5-Q10) are reported in Table~\ref{tab:subjective_full}.
For the custom questionnaire, both Intent-MPC and B-MPC significantly outperformed A-MPC across all questions, with no significant differences between Intent-MPC and B-MPC.
Analysis of the NASA-TLX subscales provides additional insight into perceived operator workload.
Both Intent-Informed MPC and B-MPC resulted in significantly lower perceived effort and frustration, as well as higher perceived performance, compared to A-MPC.
Intent-Informed MPC additionally yielded significantly lower mental demand relative to A-MPC.
Across all three control modes, physical demand and temporal demand were reported to be comparable.
Finally, no significant differences were observed between Intent-Informed MPC and B-MPC on any NASA-TLX subscale.
\section{Discussion}
We compared our approach against two MPC-based shared autonomy methods, A-MPC and B-MPC, which use distance-based arbitration strategies.
In these approaches, the level of autonomy is adjusted through discrete or proximity-triggered mechanisms rather than being continuously modulated throughout the task.
Across all objective metrics, Intent-Informed MPC significantly outperformed both baselines.
Specifically, Intent-Informed MPC achieved reductions of at least 61\% in object retrieval time, 44\% in trajectory length, and 60\% in human control input.
Subjective results further indicate that Intent-Informed MPC and B-MPC both improved perceived control, predictability, intent understanding, and responsiveness compared to A-MPC.
Overall, Intent-Informed MPC reduced or maintained cognitive workload relative to the baseline methods while improving task performance.
A-MPC struggled in our experimental task due to the close spatial arrangement of objects in the bin.
Because its formulation relies on distance-based weighting across all candidate targets, the robot may remain in full teleoperation even when near the intended object, requiring increased human input to precisely align the end-effector.
Moreover, the distance-based objective can bias the robot towards unintended targets, resulting in assistance that conflicts with the operator’s intent.
Participants frequently described this behavior as abrupt and unintuitive.
In contrast, B-MPC was perceived as more intuitive, as it operates in pure teleoperation until a predefined distance threshold is reached, after which the system autonomously approaches the goal.
However, this formulation assumes a prespecified target and does not perform intent inference, which limits its applicability to tasks where the goal is initially unknown.
Our approach, on the other hand, continuously adapts assistance based on inferred intent confidence rather than geometric proximity.
By regulating autonomy through belief entropy, the system can provide early assistance during the motion while remaining responsive to changes in operator intent.

\section{Conclusion}
In this paper, we propose an Intent-Informed Model Predictive Control (MPC) framework for adaptive shared autonomy in remote manipulation.
Unlike most shared autonomy approaches that use Model Predictive Control (MPC) primarily for constraint enforcement or trajectory correction, we employ MPC as a continuous policy blending mechanism.
Our key contribution is the use of the entropy of a belief distribution, derived from a multimodal intent inference, as an arbitration strategy to regulate control authority between the human operator and the autonomous system.
The proposed method enables continuous adaptation of autonomy levels, allowing smooth transitions between teleoperation and autonomous behavior as intent confidence changes.
Compared to baselines, the proposed method improves task performance in terms of retrieval time, trajectory length, and control input. 
\\Future work will focus on evaluating our approach on more complex manipulation tasks.
We will conduct a systematic ablation analysis to compare motion-only and multimodal (motion and gaze) intent inference.
Moreover, we will explore extending the intent inference to dynamic objects and investigating learning-based methods that utilize belief states instead of entropy for autonomy adjustment.

\bibliographystyle{IEEEtran}
\bibliography{references}

\end{document}